\def\year{2020}\relax
\documentclass[letterpaper]{article}
\usepackage{aaai20}
\usepackage{times}
\usepackage{helvet}
\usepackage{courier}
\usepackage{graphicx}
\usepackage{amsmath}
\usepackage{amssymb}
\usepackage{algorithmic}
\usepackage{algorithm}
\usepackage{array}
\usepackage{url}
\usepackage{color}
\usepackage{multirow}

\def\etal{{\emph{et al.}}}
\newcommand{\bl}[1]{\textbf{#1}}

\newcommand{\mc}[1]{\mathcal{#1}}

\begin{document}
%
\title{Exploring Reciprocal Attention \\for Salient Object Detection by Cooperative Learning}
\author{
{\normalsize Changqun Xia$^{1}$, Jia Li$^{2,1*}$, Jinming Su$^{2}$, Yonghong Tian$^{3,1*}$}\\
{\normalsize $^{1}$Peng Cheng Laboratory, Shenzhen, China} \\
{\normalsize $^{2}$State Key Laboratory of Virtual Reality Technology and Systems, SCSE, Beihang University}\\
{\normalsize $^{3}$National Engineering Laboratory for Video Technology, School of EE\&CS, Peking University}\\
\small xiachq@pcl.ac.cn, \{jiali,sujm\}@buaa.edu.cn, yhtian@pku.edu.cn\\
}

\maketitle
\begin{abstract}
Typically, objects with the same semantics are not always prominent in images containing different backgrounds. Motivated by this observation that accurately salient object detection is related to both foreground and background, we proposed a novel cooperative attention mechanism that jointly considers reciprocal relationships between background and foreground for efficient salient object detection. Concretely, we first aggregate the features at each side-out of traditional dilated FCN to extract the initial foreground and background local responses respectively. Then taking these responses as input, reciprocal attention module adaptively models the non-local dependencies between any two pixels of the foreground and background features, which is then aggregated with local features in a mutual reinforced way so as to enhance each branch to generate more discriminative foreground and background saliency map. Besides, cooperative losses are particularly designed to guide the multi-task learning of foreground and background branches, which encourages our network to obtain more complementary predictions with clear boundaries. At last, a simple but effective fusion strategy is utilized to produce the final saliency map. Comprehensive experimental results on five benchmark datasets demonstrate that our proposed method performs favorably against the state-of-the-art approaches in terms of all compared evaluation metrics.

\end{abstract}

\section{Introduction}

Salient object detection (SOD) usually aimss to detect only the most salient objects in a scene and segments the whole extent of those objects accurately.
Many fields in computer vision and image processing can be enhanced by employing saliency detection, such as content-aware image editing~\cite{cheng2010repfinder}, visual tracking~\cite{borji2012adaptive}, person re-identification~\cite{bi2014person,wu2018deep} and image retrieval~\cite{cheng2017intelligent}. Although numerous models have been proposed to detect salient objects based on different handcrafted features~\cite{xia2017and,jiang2013salient,tong2015salient}, it is still a huge challenge to detect salient objects in complex scenarios.


With a recent development of the convolutional neural network (CNNs), which intelligently learn effective feature representation, a lot of great SOD works such as~\cite{wang2017stagewise,chen2017look,zhang2017amulet} have obtained promising results on the benchmarks. Specifically, ~\cite{chen2018reverse} proposes a reverse attention to guide side-out residual learning for saliency refinement.~\cite{wang2018detect} gather contextual information for refining the convolutional features iteratively with a recurrent mechanism from the global and local view. In~\cite{zhang2018bi}, a gated bi-directional message passing module to adaptively and effectively incorporate multi-level convolutional features. And~\cite{zhang2018progressive} proposed a novel progressive attention guided module which selectively integrates multiple contextual information of multi-level features. However, these methods mainly focus on how to better integrate high-level and low-level features or multi-scale contextual information by various ways. They have not explored in depth how the proposed networks reflect the underlying essence of salient object detection, making these methods look very similar to the framework of generic object detection tasks.

\begin{figure}[t]
\centering
\includegraphics[width=1\columnwidth]{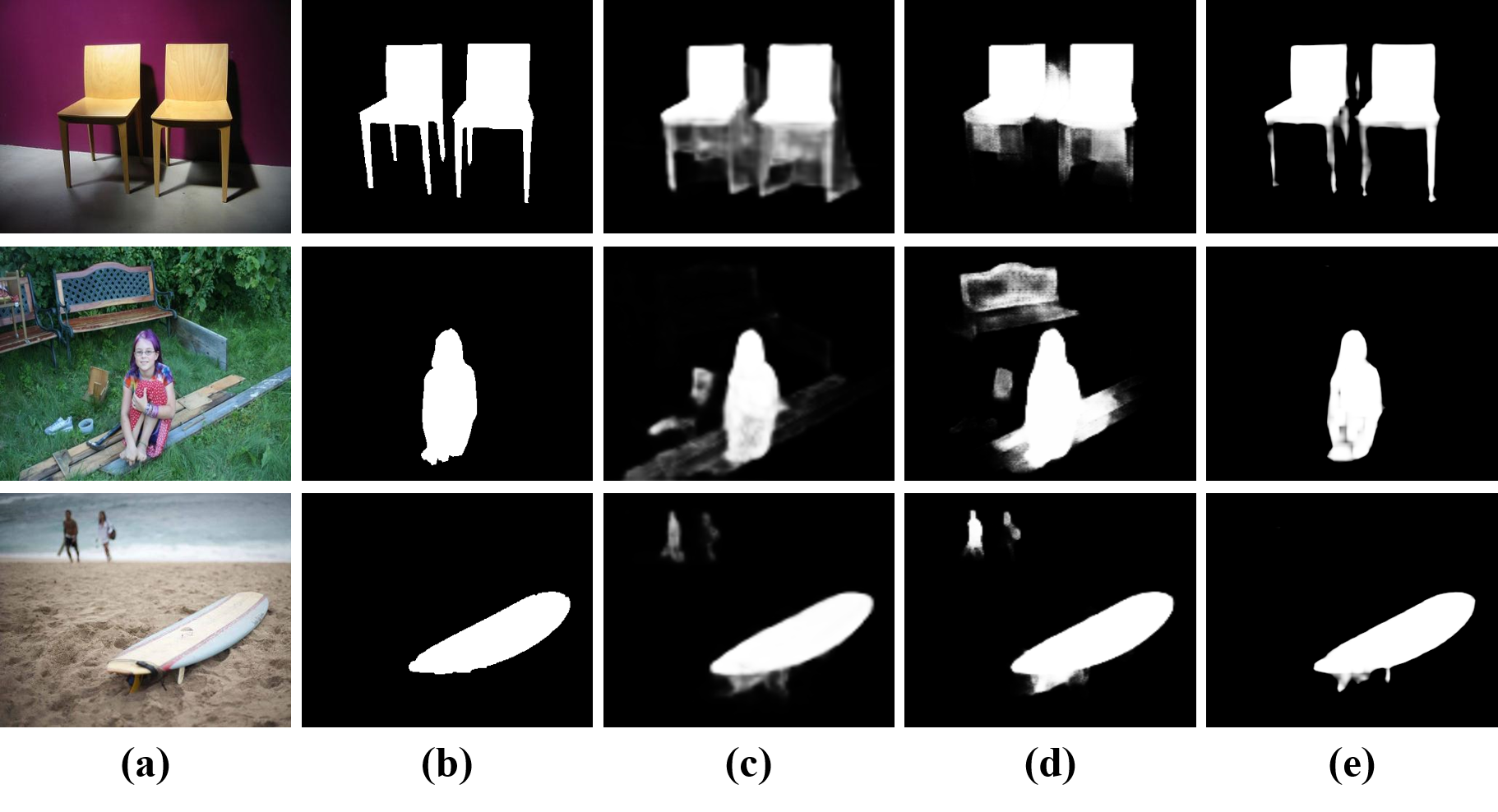}
\caption{The objects with same semantics are not always prominent in images with different backgrounds. (a)~Images, (b)~ground-truth, (c)~results of \cite{liu2018picanet}, (d)~results of \cite{luo2017non}, (e)~results of our approach. The object `Chair' in the 1st row is salient but becomes inconspicuous in the 2nd row. As the background changes, the salient object `Person' in the 2rd row is not as prominent as the object `Boat' in the 3th row. }
\label{fig:motivation}
\end{figure}

\begin{figure*}[t]
\centering
\includegraphics[width=1\textwidth,height=7.5cm]{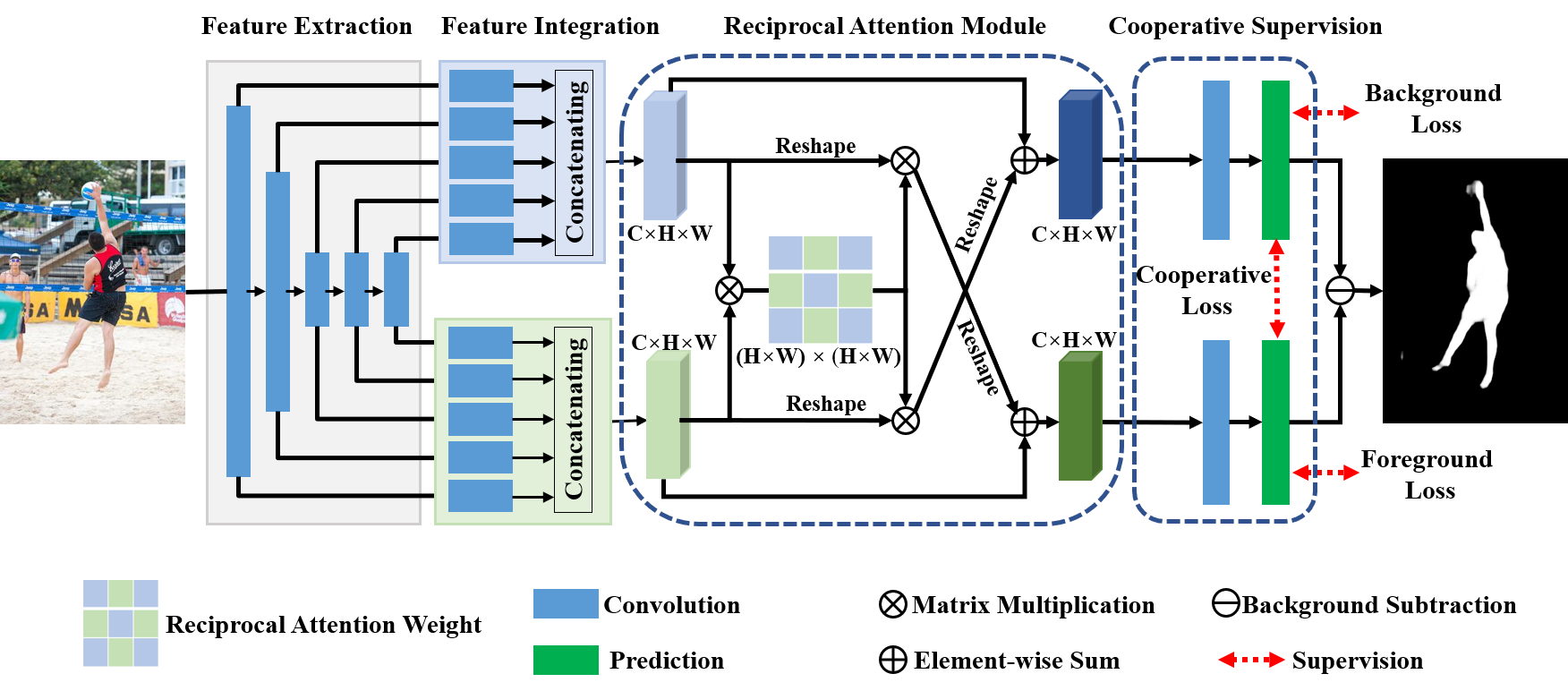}
\caption{Framework of the proposed approach (Best viewed in color).}
\label{fig:framework}
\end{figure*}


In fact, objects with the same semantics show different degrees of visual attention in images with diverse backgrounds. As shown in Fig.~\ref{fig:motivation}, the object `Chair' in the 1st row is obviously remarkable. However, due to the changes of the relative relationship between the foreground and the background of the image in the 2rd row, the salient object become the little girl. Similarly, in the 3th row, the `ship' becomes the salient object rather than the `Person'. According to the above observation, we found that reciprocal relationship between foreground and background is a key aspect in recognising what leads us to distinguish certain objects from others, which will be helpful to develop better SOD models.

Inspired by this finding, we rethink saliency detection task from the perspective of cooperative learning between foreground and background and propose a novel reciprocal attention network (RecNet) for salient object detection. As shown in Fig.~\ref{fig:framework}, the network first extracts common visual features from each side-out and then delivers them into feature integration block to obtain initial foreground and background local responses, respectively. Then two parallel features from the feature integration block would be fed into reciprocal attention module. Through matrix multiplication operation, we obtain reciprocal attention weight, which models the spatial interdependencies between any two positions between the background and foreground feature maps. Then we update the initial foreground and background features via aggregating the specified features at all positions with the reciprocal attention weight with a residual connection. As a result, reciprocal attention module can make both features contribute mutual improvement regardless of distance in spatial dimension in a mutual reinforced way. Besides, expect for the conventional supervised cross entropy loss that is utilized to train the foreground and background branches, we design cooperative loss so that the branch prediction results are more complementary and the boundary is clear. The final saliency map is obtained by a very simple but effective background substraction strategy. Experimental results on five public benchmark datasets show that our approach outperforms 12 state-of-the-art SOD models.

The main contributions of this paper are summarized as follows:
1)~We revisit the problem of SOD from the new perspective of cooperative learning between foreground and background. Compared with previous works, this scheme will be more consistent with the essence of saliency detection, which may be to helpful to develop new models.
2)~We propose a novel attention module to deal with reciprocal relationship that captures the global feature interdependencies in terms of foreground and background. In this way, the discriminative power of the foreground and background features would be enhanced.
3)~According to the characteristic of foreground and background, we design cooperative loss to encourages our network to generate more complementary predictions with clear boundaries.
4)~We conduct comprehensive experiments on five challenging datasets and achieve superior performance over the state-of-the-art approaches on all these datasets.
\section{Related Work}

Salient object detection aims to highlight salient object regions. In the past two decades, numerous models have been proposed that utilize low level features and prior cues such as intensity, color and texture~\cite{cheng2015global,xia2017and,yan2013hierarchical}.
Although these approaches can generate accurate saliency maps in most simple images, they are unable to deal with complex images due to the lack of semantic knowledge.

Recently, deep learning based approaches, in particular the convolutional neural networks (CNNs), have delivered remarkable performance in salient object detection tasks. Wang \etal~\cite{wang2015deep} first propose two deep neural networks to integrate local pixel estimation and global proposal search for salient object detection.
Li and Yu \cite{li2015visual} adopt CNNs to extract multiscale contextual features on multiscale image regions to infer saliency for each pixel and each superpixel, respectively. Similarly, Zhao \etal \cite{zhao2015saliency} use CNNs on multiple contexts to capture object saliency. In \cite{li2016deep}, an FCN based saliency model and a multiscale image region based saliency model are combined. Wang \etal \cite{kuen2016recurrent} recurrently adopt an FCN to refine saliency maps progressively. Luo \etal \cite{luo2017non} and Zhang \etal \cite{zhang2017amulet} also utilize U-Net based models to incorporate multi-level contexts to detect salient objects. Wang \etal \cite{wang2017stagewise} also use several stages to progressively refine saliency maps by combining local and global context information. In \cite{hou2017deeply}, short connections are introduced into the multi-scale side outputs within the HED network \cite{xie2015holistically} to improve saliency detection performance. Hu \etal \cite{hu2017deep} propose to adopt a level sets based loss to train their saliency detection network and use guided super-pixel filtering to refine saliency maps. Chen \etal \cite{chen2017look} propose novel two-stream fixation-semantic CNNs that can effectively detect salient objects in images. Chen \etal \cite{chen2018eccv} propose a reverse attention based side-output residual learning approach guides the whole network to sequentially discover complement object regions. Li \etal \cite{li2018contour} introduce a novel method to borrow contour knowledge for salient object detection to bridge the gap between contours and salient objects regions. Liu \etal \cite{liu2018picanet} learn to selectively attend to informative context locations for each pixel. In this way, the attended global context and multiscale local contexts can be used to effectively improve saliency detection performance. Zhang \etal \cite{zhang2018progressive} propose a novel progressive attention guided module which selectively integrates multiple contextural information of multi-level features. Zhang \etal \cite{zhang2018bi} design a gated bi-directional message passing module to integrate multi-level features, in which features from different levels adaptively pass messages to each other.
In general, these fully-supervised CNN-based methods are mainly designed to adaptively and effectively incorporate multi-level or multi-scale convolutional features, thereby  achieving good performance even when handling complex scenes.

Although these methods can obtain good results, they have three shortcomings.
1) These solutions are not closely associated with the definition of salient object. Except for the supervision using salient object training datasets, there is even no obvious difference between some of them and generic object detection task.
2) The relationship between the various cues introduced into the saliency task is established by local neighborhood convolutional or recurrent operations where the long-range dependencies are not included.
3) The training supervision of each branch is independent, lacking of the cooperative relation with each other.
Therefore, we solve the SOD task from a new perspective of cooperating learning. In this work, we propose a novel attention mechanism that jointly considers reciprocal relationships between background and foreground for efficient salient object detection.

\section{Our Proposed Approach}
In this section, we give the details of the proposed Reciprocal Attention Network (RecNet) for salient object detection. Our proposed RecNet consists of four components: feature extraction, feature integration, reciprocal attention module and cooperative supervision. The network architecture is given in Fig.~\ref{fig:framework}. At first, we will first present our base network. Then, we will introduce reciprocal attention module which captures long-range contextual interdependencies between foreground and background. At last, we describe cooperative supervision that make the predictions of foreground and background branches more complementary with clear boundaries.

\subsection{Base Network}

\bl{Feature Extraction.} We employ a pretrained residual network with the dilated strategy as the backbone to extract common features~\cite{chen2018deeplab}. Note that we remove the downsampling operations and employ dilation convolutions in the last two ResNet blocks, thus enlarging the size of the final feature map size to 1/8 of the input image with 2048 channels. This retains more details without adding extra parameters. For the sake of simplification, the subnetworks in these five blocks are denoted as $\theta_i(\pi_i),i\in\{1,\ldots,5\}$, where $\pi_i$ is the set of parameters of $\theta_i$.

\bl{Feature Integration.} Given the common features, we use two feature integration blocks with the same structure for foreground and background, respectively. Inspired by the work of~\cite{xie2015holistically}, the block is a simple module that aggregates multi-level common features and fuses them by concatenating to obtain the initial foreground or background local features. The input of each block is the feature from $\{\theta_i(\pi_i)\}_{i=1}^5$. For the feature map of each $\theta_i(\pi_i)$, we add two convolution layers with 256 kernels of $3\times{}3$ and another convolution layers with 256 kernels of $1\times{}1$, respectively. Considering the inconsistent resolution of multi-level convolutional features, we set the stride of the first two convolutions of $\theta_1(\pi_1)$ and first convolution of $\theta_2(\pi_2)$ to 2. In this way, the output feature map size of each side-out becomes 1/8 of the input image with 256 channels. At last, after the concatenation operation, the output of each block is 1/8 of the input image with 1280 channels, which are fed into the following Reciprocal Attention Module.

\subsection{Reciprocal Attention Module}
In order to describe the reciprocal relationship between the foreground and the background in images, we propose reciprocal attention module which can capture the long-range contextual dependencies between them, thus enhancing their capability of feature representation. Next we elaborate the process to adaptively aggregate mutual contextual information between background and foreground.

\begin{figure}[t]
\centering
\includegraphics[width=1\columnwidth]{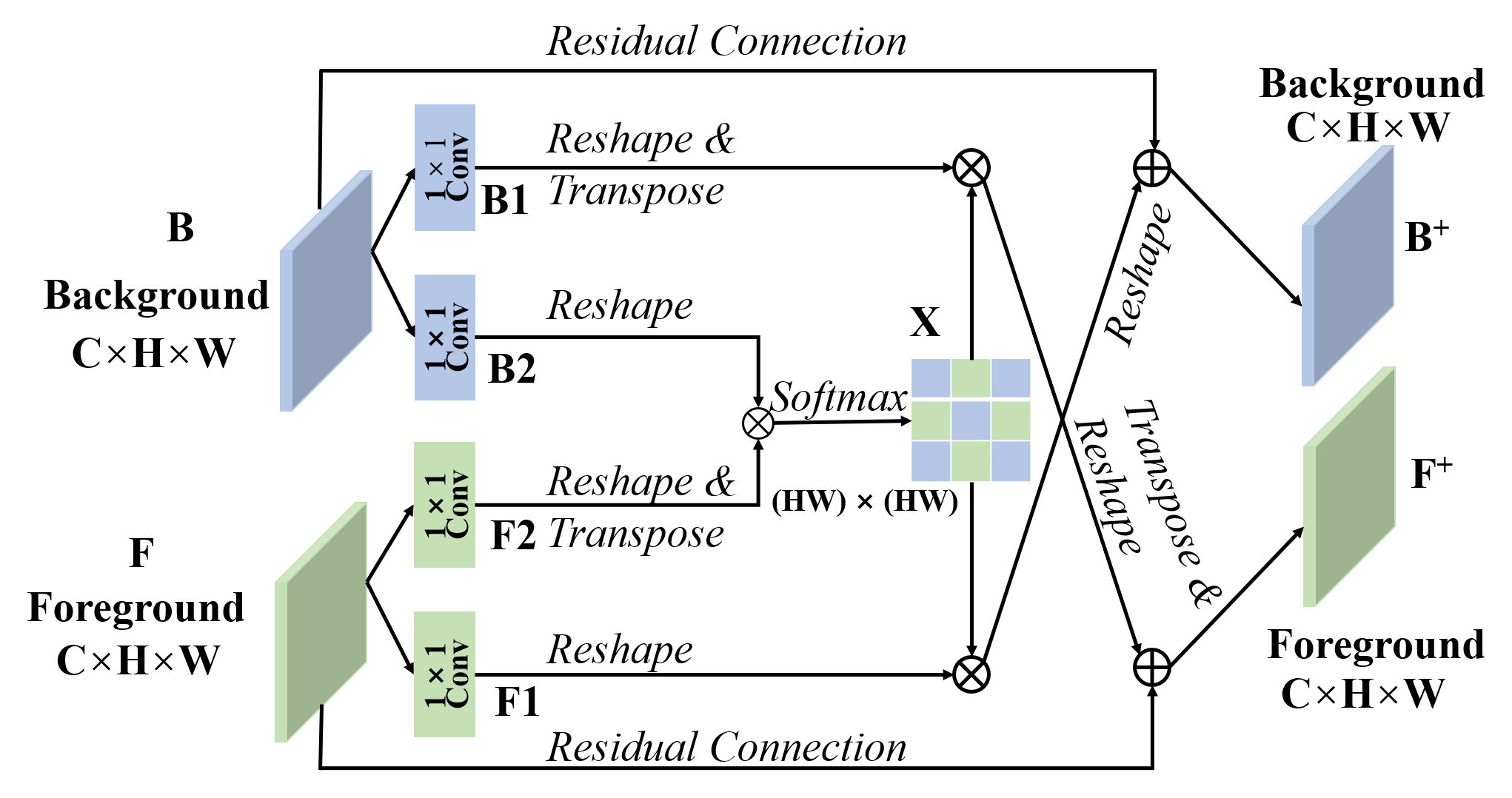}
\caption{The details of Reciprocal Attention Module between foreground (F) and background (B). (Best viewed in color)}
\label{fig:reciprocal}
\end{figure}

As illustrated in Fig.~\ref{fig:reciprocal}, given a local foreground feature~\bl{F}$\in\mathbb{R}^{C\times H\times W}$, and a local background feature~\bl{B}$\in\mathbb{R}^{C\times H\times W}$, we first feed it a convolution layers with 512 kernels of $1\times{}1$ to generate four new feature maps \bl{F1}, \bl{F2}, \bl{B1} and \bl{B2}, respectively, where~$\{\bl{F1}, \bl{F2}, \bl{B1}, \bl{B2}\}$ $\in\mathbb{R}^{C\times H\times W}$. Then we reshape them to $\in\mathbb{R}^{C\times N}$ where $N = H \times W$ is the number of features. After that we perform a matrix multiplication between the transpose of \bl{F2} and \bl{B2}, and apply a softmax layer to calculate the reciprocal attention weight map~\bl{X}$\in\mathbb{R}^{N\times N}$:

\begin{equation}\label{eq:reciprocal_weight}
x_{ij}=\frac{exp(B2_{i}\cdot F2_{j}))}{\sum_{N}^{i=1}exp(B2_{i}\cdot F2_{j}))}
\end{equation}
where $x_{ij}$ measures the $i^{th}$ position's of background feature map impact on $j^{th}$ position's of foreground feature map. Note that the more similar feature representations of any two position of background and foreground map contributes to greater correlation between them.

Meanwhile, to obtain the influence of foreground map on the background map, we first perform a matrix multiplication between \bl{F1} and~\bl{X} and reshape the result to $\in\mathbb{R}^{C\times H\times W}$. Finally, we multiply it by a scale parameter $\alpha$ and perform a element-wise sum operation with the background features \bl{B} to obtain the final output \bl{$B^{+}$} $\in\mathbb{R}^{C\times H\times W}$ as follows:

\begin{equation}\label{eq:reciprocal_background}
B^{+}_{j}=\alpha \sum_{i=1}^{N}(x_{ji}F1_{i})+B_{j}
\end{equation}
where $\alpha$ is initialized as 0 and gradually learn to assign more weight~\cite{zhang2018self}. It can be inferred from Eqs.~\eqref{eq:reciprocal_background} that the resulting background feature \bl{$B^{+}$} at each position is a weighted sum of the influence of foreground map on the background map at all positions and original background features.

Similarly, we use the same method to explore the influence of the background map on the foreground map. We perform a matrix multiplication between the transpose of \bl{B1} and~\bl{X} and reshape and transpose the corresponding result to $\in\mathbb{R}^{C\times H\times W}$. Finally, we also multiply it by a scale parameter $\beta$ and perform a element-wise sum operation with the background features \bl{F} to obtain the final output \bl{$F^{+}$}$\in\mathbb{R}^{C\times H\times W}$ as follows:

\begin{equation}\label{eq:reciprocal_foreground}
F^{+}_{j}=\beta \sum_{i=1}^{N}(x_{ji}B1_{i})+F_{j}
\end{equation}

So far, we can found the new updated foreground and background features have a global contextual view and could selectively aggregates contexts according to the reciprocal attention map. These feature representations achieve mutual gains and are more robust for salient object detection.

\subsection{Cooperative Supervision}

\bl{Cross-entropy Loss}. After the reciprocal attention module, the neural networks still split into two separate branches that address two complementary tasks, i.e., foregroundness and backgroundness estimation. Thus, we append two convolution layers with 128 kernels of $3\times{}3$ and one $1\times{}1$ kernel on top of the reciprocal attention module to output a single channel $H\times{}W$ feature map $\phi_{\mc{F}}(\pi_\mc{F})$ and $\phi_{\mc{B}}(\pi_\mc{B})$, respectively. A sigmoid layer is then used to generate corresponding foreground and background map  by minimizing the loss
\begin{equation}\label{eq:crossentropy}
L_\mc{CE}=D_{c}(Sig(\phi_{\mc{B}}(\pi_\mc{B})),G_\mc{B}) + D_{c}(Sig(\phi_{\mc{F}}(\pi_\mc{F})),G_\mc{F}),
\end{equation}
where $Sig(\cdot)$ is the sigmoid function and $D_{c}(\cdot)$ means the cross-entropy loss function.

\begin{figure}[t]
\centering
\includegraphics[width=1\columnwidth]{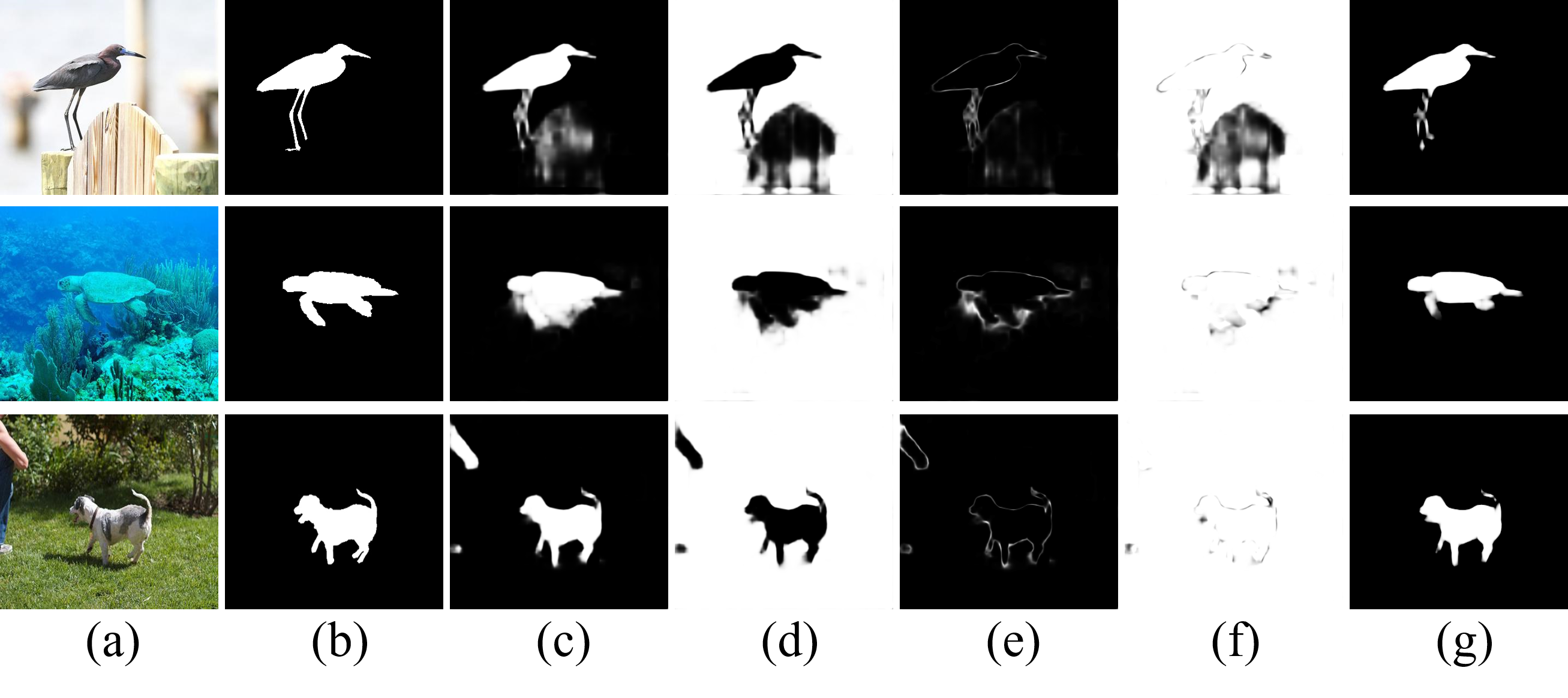}
\caption{Exploring the influence of cooperative loss. (a)~Images, (b)~ground-truth, (c) results of foreground branches w/o cooperative loss, (d)~results of background branches w/o cooperative loss, (e) union result of (c) and (d) , (f) intersection results of (c) and (d), (g)~final results with cooperative loss. The boundaries of foreground and background maps are not satisfying (see column (e)). Besides, they are not always perfectly complementary, leaving some area mistakenly predicted (see column (f)).}
\label{fig:cooperatoveloss}
\end{figure}

 As shown in Fig.~\ref{fig:cooperatoveloss}, the foreground and background maps can basically depict salient objects and distractors. However, by observing the intersection and union results of foreground and background maps, we found that the boundaries of foreground and background maps are not satisfying (see column (e)) and they are not always perfectly complementary, leaving some area mistakenly predicted (see column (f)).

 \bl{Cooperative Loss}. To reduce such error, we propose additional loss function by utilizing the idea of cooperative learning that quantifies the match of the two networks' predictions. In this way, the network is encouraged to obtain more complementary predictions with clear boundaries. Here, we use the Kullback Leibler (KL) Divergence:

 \begin{equation}\label{eq:cooperative}
 \begin{split}
L_\mc{KL}=D_{kl}(Sig(\phi_{\mc{F}}(\pi_\mc{F})),1-Sig(\phi_{\mc{B}}(\pi_\mc{B})) )  + \\
 D_{kl}(Sig(\phi_{\mc{F}}(\pi_\mc{F})).*Sig(\phi_{\mc{B}}(\pi_\mc{B})), 0)
 \end{split}
\end{equation}
where the symbol `.* represents the matrix dot product.

Thus, we use all losses above to jointly train the proposed method. As shown in column (g) in Fig.~\ref{fig:cooperatoveloss}, the final result is much more accurate.

\subsection{Training and Inference}
We use standard stochastic gradient descent algorithm to train the whole networks end-to-end using caffe. In the optimization process, the parameter of common feature extractor is initialized by the pre-trained ResNet-50 model \cite{he2016deep}, whose learning rate is set to $5 \times 10^{-9}$ with a weight decay of 0.0005 and momentum of 0.9. The learning rates of the rest layers in the proposed network are set to 10 times larger. Besides, we employ the ``poly'' learning rate policy for all experiments similar to \cite{liu2015parsenet}. The training images are not done with any special treatment except the horizontal flipping. The training process takes almost 15 hours and converges after 15k iterations with mini-batch of size 2.

During testing, the proposed network removes all the losses, and each image is resized to $384 \times 384$. After the feed-forward process, the output of the network is composed of a foreground map and a background map. We use a very simple but effective fusion strategy based on background substraction operation~\cite{piccardi2004background, zhang2017learning}, i.e.,
 \begin{equation}\label{eq:cooperative}
 \begin{split}
Sal = relu(Sig(\phi_{\mc{F}}(\pi_\mc{F})) - Sig(\phi_{\mc{B}}(\pi_\mc{B})))
 \end{split}
\end{equation}
 where $relu(\cdot)$ means rectified linear unit function.

 The proposed method runs at about 13 fps on our computer with a 3.60GHz CPU and a GTX 1080ti GPU.


\section{Experiments and Results}
\subsection{Experimental Setup}
\textbf{Datasets.}
To evaluate the performance of the proposed approach, we conduct experiments on five benchmark datasets~\cite{yan2013hierarchical,li2014secrets,li2015visual,wang2017learning,xia2017and}. Details of these datasets are described briefly as follows: ECSSD~\cite{yan2013hierarchical} contains 1,000 images with complex structures and obvious semantically meaningful objects. PASCAL-S~\cite{li2014secrets} includes 850 natural images that are pre-segmented into objects or regions and free-viewed by 8 subjects in eye-tracking tests for salient object annotation. HKU-IS~\cite{li2015visual} comprises 4,447 images and lots of images contain multiple disconnected salient objects or salient objects that touch image boundaries. DUTS~\cite{wang2017learning} is a large scale dataset containing 10533 training images (denoted as DUTS-TR) and 5019 test images(denoted as DUTS-TE). The images are challenging with salient objects of varied locations and scales as well as complex background. XPIE~\cite{xia2017and} has 10000 images covering a variety of simple and complex scenes with different numbers, sizes and positions of salient objects.

\textbf{Evaluation Metrics.}
In the comparisons, we adopt the F-measure curves, adaptive F-measure, weighted F-measure and mean absolute error (MAE) as the evaluation metrics. In computing F-measure curves,
the precision and recall are first computed by binarizing the saliency maps with a threshold sliding from 0 to 255 and compare the binary maps with ground-truth maps. At each threshold, F-measure is computed as
 \begin{align}
 \text{F-measure} = \frac{{(1 + {\beta ^2}){\rm{\cdot}}\text{Precision}{\rm{\cdot}}\text{Recall}}}{{{\beta ^2}{\rm{\cdot}}\text{Precision} + \text{Recall}}},
  \end{align}
where $\beta$ is set to 0.3 as in \cite{achanta2009frequency}. Besides, we report adaptive F-measure (F$_{\beta}$) using an adaptive threshold for generating a binary saliency map. The adaptive threshold is computed as twice the mean value of the saliency map. Meanwhile, a unified weighted F-measure (F$_{\beta}$) score is computed as in \cite{margolin2014evaluate} to reflect the overall performance. In addition, MAE is calculated as the average absolute per-pixel difference between the gray-scale saliency maps and the ground-truth saliency maps.

\subsection{Comparisons with the State-of-the-Art}
We compare our approach denoted as \textbf{RecNet} with 12 state-of-the-art methods, including ELD \cite{lee2016deep}, UCF \cite{zhang2017learning}, NLDF \cite{luo2017non}, Amulet \cite{zhang2017amulet}, FSN \cite{chen2017look}, SRM \cite{wang2017stagewise}, C2SNet~\cite{li2018contour}, RA~\cite{chen2018eccv}, Picanet~\cite{liu2018picanet}, PAGRN \cite{zhang2018progressive}, R3Net~\cite{deng2018r3net} and DGRL~\cite{wang2018detect}. For fair comparison, we obtain the saliency maps of different methods from the authors or the deployment codes provided by the authors.

\begin{table*}[t]
\centering
\caption{Performance of 12 state-of-the-arts and the proposed method on six benchmark datasets. Smaller MAE and larger F$^{w}_{\beta}$ correspond to better performance. The best three results are in {\color{red}{red}}, {\color{green}{green}} and {\color{blue}{blue}} fonts, respectively. ``-" means the result can’t be obtained.}
\setlength{\tabcolsep}{1mm}{
\renewcommand\arraystretch{1.2}
\begin{tabular}{|c|c|c|c|c|c|c|c|c|c|c|c|c|c|c|c|}
\hline
\multirow{2}*{Models} & \multicolumn{3}{|c|}{ECSSD} & \multicolumn{3}{|c|}{PASCAL-S} & \multicolumn{3}{|c|}{HKU-IS} & \multicolumn{3}{|c|}{DUTS-TE} & \multicolumn{3}{|c|}{XPIE}\\
\cline{2-16}
& MAE1 & F$^{w}_{\beta}$ & F$_{\beta}$ & MAE & F$^{w}_{\beta}$ & F$_{\beta}$
& MAE & F$^{w}_{\beta}$ & F$_{\beta}$ & MAE & F$^{w}_{\beta}$ & F$_{\beta}$ & MAE & F$^{w}_{\beta}$ & F$_{\beta}$ \\

\hline
ELD
            & 0.078 & 0.786 & 0.829  & 0.124 & 0.669 & 0.746 & 0.063 & 0.780 & 0.827 & 0.092 & 0.608 & 0.647 & 0.085 & 0.698 & 0.746 \\
UCF
            & 0.069 & 0.807 & 0.865  & 0.116 & 0.696 & 0.776 & 0.062 & 0.779 & 0.838 & 0.112 & 0.596 & 0.670 & 0.095 & 0.693 & 0.773 \\
NLDF
            & 0.063 & 0.839 & 0.892  & 0.101 & 0.737 & 0.806 & 0.048 & 0.838 & 0.884 & 0.065 & 0.710 & 0.762 & 0.068 & 0.762 & 0.825 \\
Amulet
            & 0.059 & 0.840 & 0.882  & 0.099 & 0.736 & 0.795 & 0.051 & 0.817 & 0.853 & 0.085 & 0.658 & 0.705 & 0.074 & 0.743 & 0.796 \\
FSN
            & 0.053 & 0.862 & 0.889  & 0.095 & 0.751 & 0.804 & 0.044 & 0.845 & 0.869 & 0.069 & 0.692 & 0.728 & 0.066 & 0.762 & 0.812 \\
SRM
            & 0.054 & 0.853 & 0.902  & 0.086 & 0.759 & 0.820 & 0.046 & 0.835 & 0.882 & 0.059 & 0.722 & 0.771 & 0.057 & 0.783 & 0.841 \\
C2SNet
            & 0.057 & 0.844 & 0.878  & 0.086 & 0.764 & 0.805 & 0.050 & 0.823 & 0.854 & 0.065 & 0.705 & 0.740 & 0.066 & 0.764 & 0.807 \\
RA
            & 0.056 & 0.857 & 0.901  & 0.105 & 0.734 & 0.811 & 0.045 & 0.843 & 0.881 & 0.059 & 0.740 & 0.772 & 0.067 & 0.776 & 0.836 \\
Picanet
            & 0.047 & 0.866 & 0.902  & {\color{blue}{0.077}} & 0.778 & 0.826 & 0.043 & 0.840 & 0.878 & {\color{green}{0.051}} & 0.755 & 0.778 & {\color{blue}{0.052}} & 0.799 & 0.843 \\
PAGRN
            & 0.061 & 0.834 & {\color{blue}{0.912}} & 0.094 & 0.733 & {\color{blue}{0.831}} & 0.048 & 0.820 & {\color{blue}{0.896}} & {\color{blue}{0.055}} & 0.724 & {\color{blue}{0.804}} & - & - & -\\
R3Net
            & {\color{green}{0.040}} & {\color{green}{0.902}} & {\color{green}{0.924}} & 0.095 & 0.760 & {\color{green}{0.834}} & {\color{green}{0.036}} & {\color{green}{0.877}} & {\color{green}{0.902}} & 0.057 & {\color{green}{0.765}} & {\color{green}{0.805}} & 0.058 & {\color{blue}{0.805}} & {\color{blue}{0.854}} \\
DGRL
            & {\color{blue}{0.043}} & {\color{blue}{0.883}} & 0.910 & {\color{green}{0.076}} & {\color{green}{0.788}} & 0.826 & {\color{blue}{0.037}} & {\color{blue}{0.865}} & 0.888 & {\color{green}{0.051}} & {\color{blue}{0.760}} & 0.781 & {\color{green}{0.048}} & {\color{green}{0.818}} & {\color{green}{0.859}} \\
\textbf{RecNet}
            & {\color{red}{0.035}} & {\color{red}{0.914}} & {\color{red}{0.942}} & {\color{red}{0.067}} & {\color{red}{0.815}} & {\color{red}{0.852}} & {\color{red}{0.030}} & {\color{red}{0.899}} & {\color{red}{0.925}} & {\color{red}{0.045}} & {\color{red}{0.803}} & {\color{red}{0.840}} & {\color{red}{0.046}} & {\color{red}{0.836}} & {\color{red}{0.877}} \\
\hline
\end{tabular}}
\label{tab:result-all}
\end{table*}

\begin{figure*}[t]
\centering
\includegraphics[width=1\textwidth]{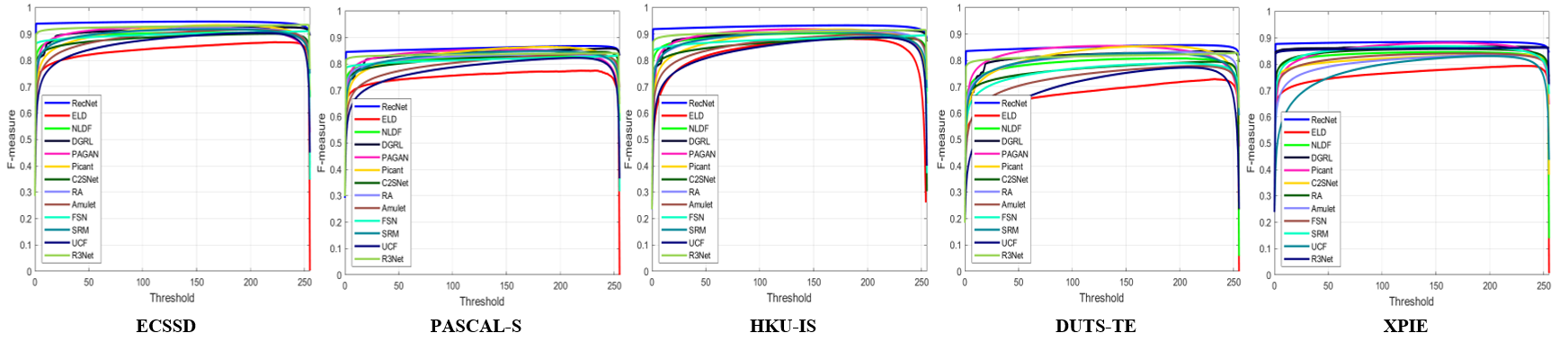}
\caption{The F-measure curves of 12 state-of-the-arts and our approach are listed across five benchmark datasets.}
\label{fig:prfmeasure}
\end{figure*}

\begin{figure*}[t]
\centering
\includegraphics[width=1\textwidth]{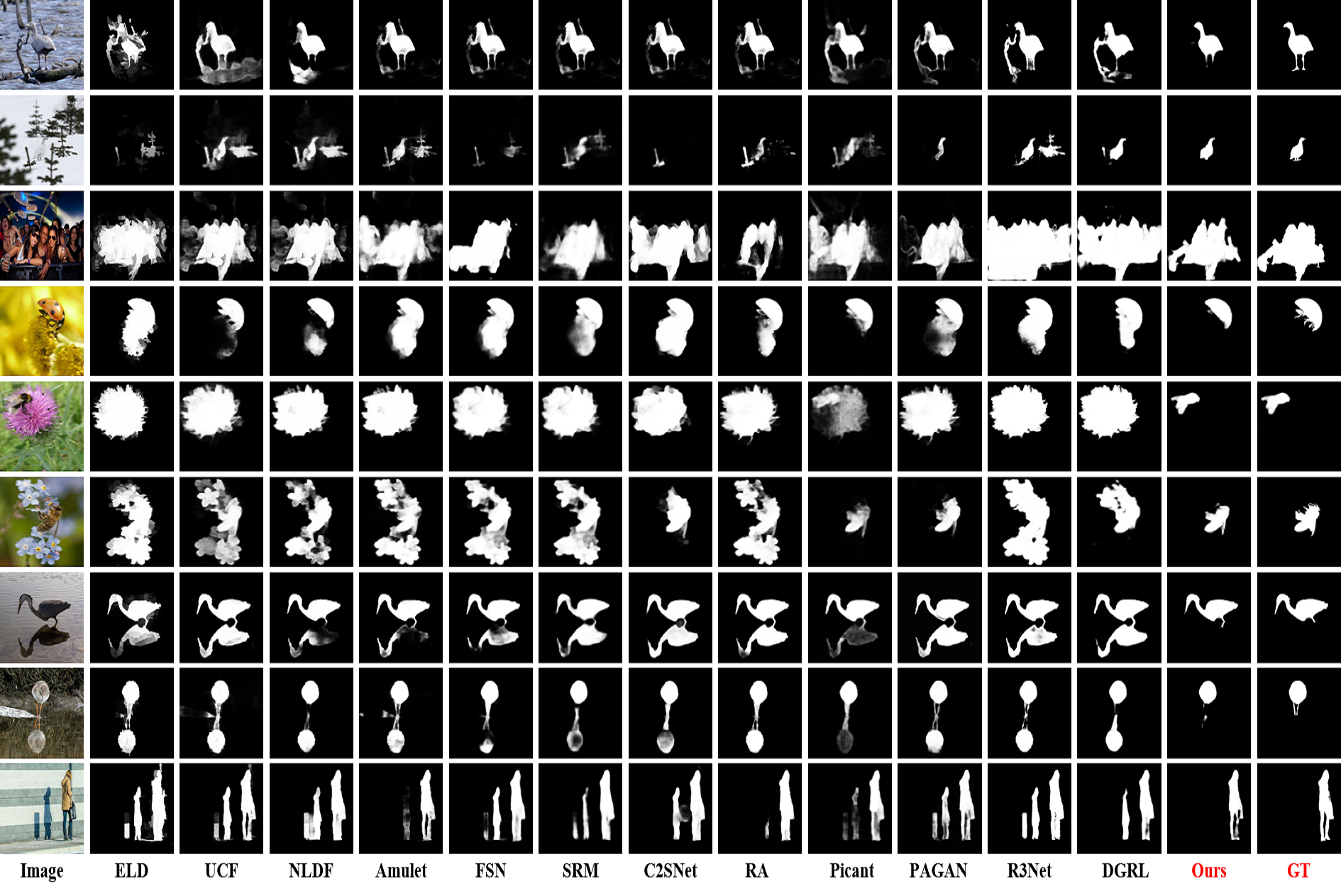}
\caption{Qualitative comparisons of th state-of-the-art algorithms and the our approach. GT means ground-truth masks of salient objects. The images are selected from five datasets for testing.}
\label{fig:result-all}
\end{figure*}

The proposed approach is compared with 12 state-of-the-art saliency detection methods on five datasets. The quantitative comparison results are shown in Tab.\ref{tab:result-all} and Fig.~\ref{fig:prfmeasure}. From Tab.\ref{tab:result-all}, we can see that our approach, RecNet, consistently outperforms all the other 12 approaches on all the five datasets in terms of all compared evaluation metrics. It is worth noting that the F$_{\beta}$ score of our method is significantly improved compared with the second best results on HKU-IS and DUTS-TE, 0.925 against 0.902 and 0.840 against 0.805.

Fig.~\ref{fig:result-all} show example saliency maps generated by our approach as well as other 12 state-of-the-art methods. We can see that salient objects can pop-out as a whole with clear boundaries by the proposed method. We can find that many methods fail to detect the salient objects with large changed appearance as a whole as depicted in the row of 1 to 3. These observations indicated the cooperative loss are important to deal with integrity of objects and clarity of boundaries for SOD. In addition, when salient objects share the same attributes (such as locations) with background, the background is very easy to be mistakenly detected by many methods, as shown in the row of 4 to 6. In our approach, the relative relationship between background and foreground is guaranteed by the novel reciprocal attention module, which will can capture the long-range contextual dependencies. Moreover, three extra examples about more difficult scenes are shown in the last three rows of Fig. \ref{fig:result-all}, our methods still obtain the impressive results with accurate salient object localization.

By observing the results illustrated in Fig.~\ref{fig:result-all}, we find that the success of our approach in such comparisons can be explained from three perspectives. First, we adopt a new way to introduce background cues. Compared with the prior methods, most of them used background as a prior to generate salient seeds. In our work, the background and foreground are jointly to detect salient objects by collaborative learning, which is more consistent with the definition of salient objects. Second, we explore the novel mechanism of reciprocal attention to capture the long range dependency between foreground and background. In contrast to the progressive behavior of recurrent and convolutional operations, the proposed reciprocal attention module can directly compute the interaction between any two positions of background and foreground maps. 3) The design of additional cooperative loss can make the prediction of the two network branches more complementary and its boundary more clear.

\begin{table*}[t]
\centering
\caption{Performance of the four different setting of the proposed approach on five benchmark datasets.}
\setlength{\tabcolsep}{0.38mm}{
\renewcommand\arraystretch{1.05}
\begin{tabular}{|c|c|c|c|c|c|c|c|c|c|c|c|c|c|c|c|}
\hline
\multirow{2}*{Models} & \multicolumn{3}{|c|}{ECSSD} & \multicolumn{3}{|c|}{PASCAL-S} & \multicolumn{3}{|c|}{HKU-IS} & \multicolumn{3}{|c|}{DUTS-TE} & \multicolumn{3}{|c|}{XPIE}\\
\cline{2-16}
& MAE & F$^{w}_{\beta}$ & F$_{\beta}$ & MAE & F$^{w}_{\beta}$ & F$_{\beta}$
& MAE & F$^{w}_{\beta}$ & F$_{\beta}$ & MAE & F$^{w}_{\beta}$ & F$_{\beta}$ & MAE & F$^{w}_{\beta}$ & F$_{\beta}$ \\

\hline
Backbone+Foreground
            & 0.040 & 0.893 & 0.914 & 0.072 & 0.785 & 0.817 & 0.034 & 0.874 & 0.890 & 0.053 & 0.753 & 0.774 & 0.050 & 0.811 & 0.843  \\
Backbone+Two Branches
            & 0.039 & 0.904 & 0.933 & 0.069 & 0.807 & 0.845 & 0.032 & 0.890 & 0.918 & 0.048 & 0.788 & 0.827 & 0.049 & 0.826 & 0.869 \\
Background-RAM
            & 0.038 & 0.907 & 0.936 & 0.068 & 0.809 & 0.845 & 0.032 & 0.892 & 0.918 & 0.047 & 0.796 & 0.831 & 0.048 & 0.831 & 0.873  \\
Foreground-RAM
            & 0.036 & 0.910 & 0.935 & 0.069 & 0.809 & 0.843 & 0.032 & 0.890 & 0.914 & 0.046 & 0.801 & 0.831 & 0.047 & 0.835 & 0.873 \\
RecNet w/o cooperative loss
            & 0.036 & 0.912 & 0.937 & 0.067 & 0.810 & 0.847 & 0.031 & 0.893 & 0.920 & 0.047 & 0.799 & 0.837 & 0.047 & 0.832 & 0.875 \\
\textbf{RecNet}
            & \textbf{0.035} & \textbf{0.914} & \textbf{0.942} & \textbf{0.067} & \textbf{0.815} & \textbf{0.852} & \textbf{0.030} & \textbf{0.899} & \textbf{0.925} & \textbf{0.045} & \textbf{0.803} & \textbf{0.840} & \textbf{0.046} & \textbf{0.836} & \textbf{0.877} \\
\hline
\end{tabular}}
\label{tab:comparisons-of-four-reciprocal}
\end{table*}

\subsection{Ablation Analysis}
To validate the effectiveness of different components of the proposed method, we conduct several experiments on all the five datasets to compare the performance variations of our methods with different experimental settings.

\textbf{Effectiveness of Reciprocal Attention Module.}
To investigate the efficacy of the proposed reciprocal attention module (RAM), we conduct ablation experiments across all five datasets by introducing four different settings for comparisons. The first setting denoted as `Backbone+Foreground' means we add 3 convolution layers directly appending on the top of  feature aggregation. Note that there is only one foreground branch in this setting. The second setting denoted as `Backbone+Two Branches'. In this setting, the network has two branches, predicting the foreground and background respectively. To explore the influenceof the reciprocal attention module, we define two setting additionally. In the setting `Background-RAM', Eqs.~\eqref{eq:reciprocal_foreground} is disabled and Eqs.~\eqref{eq:reciprocal_background} doesnot work in `Foreground-RAM'. Note that our proposed approach RecNet combines two branches, RAM and cooperative loss.

For a comprehensive comparison, MAE, F$^{w}_{\beta}$ and F$_{\beta}$ scores of above-mentioned setting and our approach RecNet are evaluated on five benchmark datasets. The comparison results are listed in Tab.\ref{tab:comparisons-of-four-reciprocal}. We can observe that compared with setting 1, the performance of setting 2 is greatly improved due to the introduction of background cues. This indicates that background cues provides a strong cooperative effect in salient object detection. In settings 3 and 4, although there is only one-way dependency in RAM, performance can still be improved. While reciprocal attemtion module in Recnet has a complete ability to capture the relationship between foreground and background, it leads a remarkable improvement of performance.

\begin{table*}[t]
\centering
\caption{Performance of different fusion setting of the proposed approach on five benchmark datasets.}
\setlength{\tabcolsep}{0.38mm}{
\renewcommand\arraystretch{1.05}
\begin{tabular}{|c|c|c|c|c|c|c|c|c|c|c|c|c|c|c|c|}
\hline
\multirow{2}*{Models} & \multicolumn{3}{|c|}{ECSSD} & \multicolumn{3}{|c|}{PASCAL-S} & \multicolumn{3}{|c|}{HKU-IS} & \multicolumn{3}{|c|}{DUTS-TE} & \multicolumn{3}{|c|}{XPIE}\\
\cline{2-16}
& MAE & F$^{w}_{\beta}$ & F$_{\beta}$ & MAE & F$^{w}_{\beta}$ & F$_{\beta}$
& MAE & F$^{w}_{\beta}$ & F$_{\beta}$ & MAE & F$^{w}_{\beta}$ & F$_{\beta}$ & MAE & F$^{w}_{\beta}$ & F$_{\beta}$ \\

\hline
Foreground
            & 0.036 & 0.903 & 0.921 & 0.071 & 0.794 & 0.827 & 0.032 & 0.881 & 0.895 & 0.052 & 0.769 & 0.790 & 0.049 & 0.821 & 0.853  \\
1-Background
            & 0.036 & 0.905 & 0.922 & 0.072 & 0.791 & 0.823 & 0.032 & 0.879 & 0.892 & 0.054 & 0.764 & 0.784 & 0.050 & 0.818 & 0.849 \\
Foreground+1-Background
            & 0.041 & 0.874 & 0.919 & 0.078 & 0.759 & 0.824 & 0.038 & 0.847 & 0.894 & 0.058 & 0.726 & 0.787 & 0.056 & 0.776 & 0.851  \\
\textbf{RecNet}
            & \textbf{0.035} & \textbf{0.914} & \textbf{0.942} & \textbf{0.067} & \textbf{0.815} & \textbf{0.852} & \textbf{0.030} & \textbf{0.899} & \textbf{0.925} & \textbf{0.045} & \textbf{0.803} & \textbf{0.840} & \textbf{0.046} & \textbf{0.836} & \textbf{0.877} \\
\hline
\end{tabular}}
\label{tab:comparisons-of-fusion}
\end{table*}

\textbf{Effectiveness of Cooperative Loss.}
In addition to use cross entropy loss function in network training, we also propose cooperative loss to improve the discrimination ability of prediction results. Therefore, we will investigate the impact of cooperative loss after removing it.
As shown in Tab.\ref{tab:comparisons-of-four-reciprocal}, we can see that after the cooperative loss is removed, the performance of each dataset decreases to different degrees, which shows that the proposed loss is effective.

\textbf{Effectiveness of Fusion Strategy.}
In this work, we use background subtraction operation to fuse the foreground map the background map. Here we also analyze the impact of different fusion strategies on performance. We have designed 3 different schemes. The first is to directly use the result of foreground prediction as the final result. The second is to use the inversion of background map as the result. The third method is to average the results of the first and second scheme.  The comparison of these three schemes and our RecNet is listed in Tab.\ref{tab:comparisons-of-fusion}. From Tab.\ref{tab:comparisons-of-fusion}, we find that the best performance is achieved by adopting the background substraction way. This substraction strategy not only increase the pixel-level discrimination but also captures context contrast information.

\section{Conclusion}
In this paper, we revisit the problem of SOD from the perspective of cooperative learning of background and foreground. Compared with previous work, this scheme will be more consistent with the essence of saliency detection, which may be to helpful to develop new models. To solve this problem, we propose a novel attention module to deal with reciprocal relationships that captures the global feature interdependencies in terms of foreground and background. In this way, the feature discriminative power is enhanced to help perceiving and localizing foreground objects and background distractors. Moreover, we also propose cooperative loss to encourage our network to obtain more complementary predictions with clear boundaries. Extensive experiments on five benchmark datasets have validated the effectiveness of the proposed approach.

\bibliographystyle{aaai}
\bibliography{RefRecNet}
\end{document}